\documentclass[letterpaper, 10 pt, conference]{ieeeconf}  

\IEEEoverridecommandlockouts                              

\usepackage{graphics} 
\usepackage{graphicx}
\usepackage{graphicx}
\usepackage{epsfig} 
\usepackage{mathptmx} 
\usepackage{times} 
\usepackage{amsmath} 
\usepackage{amssymb}  
\usepackage{amsmath}
\usepackage{booktabs}
\usepackage{microtype}

\usepackage{amsmath,amsfonts}
\usepackage{algorithmic}
\usepackage{algorithm}
\usepackage{float}

\title{\LARGE \bf
Trajectory Planning of a Curtain Wall Installation Robot Based on Biomimetic Mechanisms*
}

\author{Xiao Liu$^{1}$, Weijun Wang$^{1}$, Tianlun Huang$^{1}$, Zhiyong Wang$^{1}$, Wei Feng $^{1,2*}$
\thanks{*corresponding author. e-mail: wei.feng@siat.ac.cn}
\thanks{$^{1}$ All authors are with Shenzhen Institute of Advanced Technology, Chinese Academy of Sciences, Shenzhen, 518055, China.
        {\tt\small  Contact: xiao.liu1@siat.ac.cn}}%
 \thanks{$^{2}$ All authors are with Shenzhen University of Advanced Technology and University of Chinese Academy of Sciences, Shenzhen, 518055, China.
        {\tt\small  Contact: wei.feng@siat.ac.cn}}%
}

\begin{document}

\maketitle
\thispagestyle{empty}
\pagestyle{empty}

\begin{abstract}
As the robotics market rapidly evolves, energy consumption has become a critical issue, particularly restricting the application of construction robots. To tackle this challenge, our study innovatively draws inspiration from the mechanics of human upper limb movements during weight lifting, proposing a bio-inspired trajectory planning framework that incorporates human energy conversion principles. By collecting motion trajectories and electromyography (EMG) signals during dumbbell curls, we construct an anthropomorphic trajectory planning that integrates human force exertion patterns and energy consumption patterns. Utilizing the Particle Swarm Optimization (PSO) algorithm, we achieve dynamic load distribution for robotic arm trajectory planning based on human-like movement features. In practical application, these bio-inspired movement characteristics are applied to curtain wall installation tasks, validating the correctness and superiority of our trajectory planning method. Simulation results demonstrate a 48.4\% reduction in energy consumption through intelligent conversion between kinetic and potential energy. This approach provides new insights and theoretical support for optimizing energy use in curtain wall installation robots during actual handling tasks.
\end{abstract}

\section{INTRODUCTION}

In recent years, biomimetics has become a focus for researchers worldwide. Biomimetics is a multidisciplinary field that studies the structure, shape, and principles of biological systems, and the interactions between them, providing researchers with new design inspiration and concepts [1]-[3]. However, current research faces issues such as insufficient exploration of biological movement mechanisms, traditional approaches to shape design, material selection, driving and control methods, and suboptimal energy utilization, resulting in designs that are "similar in form but lacking in essence," falling far short of practical applications [4]-[6].

Studies on human upper limb movements are mostly limited to human-machine interaction and motion simulation, such as upper limb exoskeleton robots [7]-[9], and bionic arms [10]-[12] that achieve human-like movements, focusing primarily on structural similarity without an in-depth exploration of the movement mechanism itself.
\begin{figure}[htb]
      \centering
      \includegraphics[width=\columnwidth]{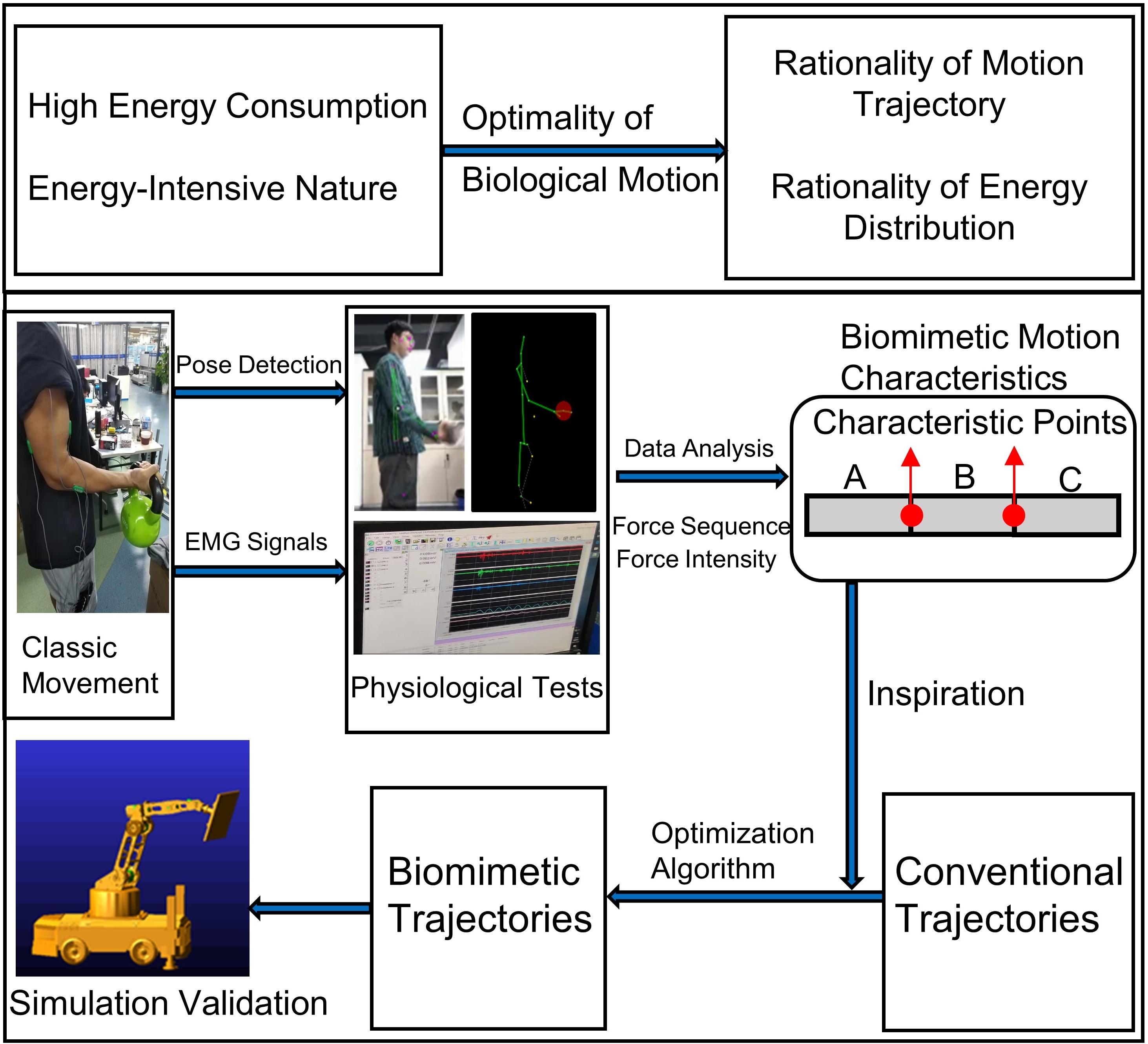}
      \caption{Overview of the Proposed Method}
      \label{Figure 1}
   \end{figure}
Both conscious and unconscious movements of the human upper limb can be considered coordinated movements controlled by the brain and high-level decision-making systems [13]. Understanding the control mechanisms, control parameters, indices, and control strategies of coordinated human upper limb movements is of tremendous value for robot motion planning and intelligent control. Commonly used parameters to describe human upper limb movement patterns include joint angles, angular velocity, electromyography (EMG) signals, mobility indices, and performance indices [14]. [15] investigated the relationship between average task completion time, upper limb movement speed, acceleration, movement amplitude, and target size based on information theory. [16] proposed the equilibrium point control theory, suggesting that upper limb movement involves a balance of muscle force and external force in the time domain, with the human body constantly searching for equilibrium points to achieve coordinated movements. [17] proposed a motion planning system based on "learning through demonstration" for upper limb exoskeletons, allowing successful assistance of patients in daily life activities (ADLs) in unstructured environments while ensuring compliance with humanoid standards throughout the entire human-machine workspace. [18] proposed a computational framework with human movement characteristics, mimicking the mechanisms that control and realize human upper limb movements, achieving smooth trajectories and natural obstacle-avoidance capabilities in humanoid motion. [19] studied a method for accurately mapping human arm movements using sEMG signals, determining arm movements through sEMG signals. [20] estimated upper limb joint torque through an electromyography-driven model.

Current bionic robotics research primarily focuses on mimicking external forms or basic motion trajectories, lacking an in-depth analysis of energy conversion mechanisms in human movements. Significant limitations remain in core aspects like energy optimization and load regulation, resulting in superficial biomimicry without functional depth. Existing methods commonly exhibit three main deficiencies:(1)Single-dimensional Data: Reliance on motion capture systems for spatial trajectories without concurrent EMG signals and joint torques, leading to incomplete energy transfer pathway modeling. (2)Algorithmic Rigidity: Application of generic algorithms (e.g., genetic algorithms, particle swarm optimization)  without incorporating biological motion characteristics as constraints. (3)Lack of Practical Applications: Focus solely on human imitation and bionics without application to specific engineering fields. 

In contrast, our study advances traditional bionic principles by deeply analyzing the mechanics of human upper limb movements during heavy lifting. We introduce, for the first time, the principle of kinetic-potential energy conversion into robotic trajectory planning, addressing issues of insufficient energy transfer and uneven load distribution. By synthesizing human movement coordination rules and optimality features and adopting a research methodology that combines theoretical research, numerical simulation, and simulation experiments, this study guides the curtain wall installation process, providing new insights and theoretical foundations for bio-inspired trajectory planning. 

\section{Biomimetic mechanism}

\subsection{System block diagram}

The dumbbell curl is a typical upper-limb functional movement, characterized by a highly standardized motion pattern, relatively simple joint involvement, and distinct muscle activation features. It primarily involves the coordinated motion of the shoulder and elbow joints, and is performed by the combined action of multiple muscles, including the biceps brachii and deltoid. During the execution of the dumbbell curl, the process of force transmission and distribution effectively reflects the human body's strategies for regulating energy efficiency during dynamic tasks. Compared with other movements, the dumbbell curl more clearly demonstrates the process of energy conversion and better reflects the force distribution patterns under different postures. It allows for an in-depth investigation into how the human body adjusts the timing and intensity of muscle activation to achieve more efficient energy utilization.Given its high repeatability and physiological representativeness, this study selected the dumbbell curl as the demonstration task for human subject experiments, to support subsequent electromyography (EMG) signal acquisition, kinematic data collection, and related experimental analyses.

The system's overall structure and the interrelationships among its components are depicted in Figure 1, providing an overview of the design principles and workflow. We collected EMG signals during dumbbell curl exercises using surface EMG sensors (Datalog) and combined this data with posture detection from Mediapipe to establish a comprehensive motion-EMG coupling analysis framework. By analyzing energy conversion patterns, we identified key characteristic points in the movement. These points were integrated into the motion trajectory using Particle Swarm Optimization (PSO).

\subsection{Modeling of Human Upper Limb Movement}
Human models are crucial for motion analysis. This study focuses on analyzing energy conversion mechanisms in upper limb movements rather than full biomechanical modeling. To minimize joint coupling interference, we simplified the upper limb to a 2-degree-of-freedom (DOF) planar stick model within the sagittal plane, focusing on flexion-extension movements of the shoulder and elbow joints while ignoring hand and wrist motions. This simplification preserves the essential characteristics of shoulder-elbow coordination, enabling clearer observation of energy conversion patterns. It also significantly reduces the degrees of freedom, lowering system complexity and enhancing computational efficiency. As shown in Figure 2, the shoulder joint is fixed at the coordinate origin, the upper arm is equivalent to link 1, and the forearm and hand are merged into link 2, with the two links connected by the elbow joint. During the process of the human upper limb arm performing a bicep curl with a barbell, the biceps brachii and the brachialis muscles act as the main force-generating muscles and are known as the prime movers. Through anatomical analysis and research, it is more convenient and accurate to study the bicep curl movement by equivalently modeling the upper arm and lower arm as rigid bodies. On the basis of the rigid model, the main body and related ideal rotary joints are formed. 

\begin{figure}[htb]
      \centering
      \includegraphics[width=\columnwidth]{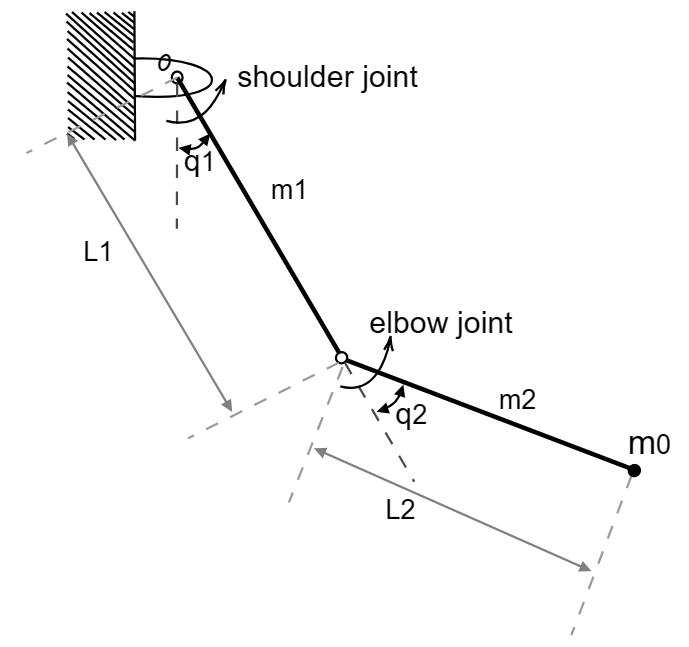}
      \caption{Simplified upper limb model}
      \label{Figure 2}
   \end{figure}

Establish a basic dynamic model, where \( \tau \) represents the joint torque. The dynamics model is constructed using the Lagrangian method, which is an energy-based dynamic approach.

Link 1:

\begin{equation}
T_{1p} = \frac{1}{2} m v_{1}^{2} = \frac{m l^{2}}{8} 
\end{equation}
The rotational kinetic energy around itself is:
\begin{equation}
\boldsymbol{T}_{1r}=\frac12\boldsymbol{J}\boldsymbol{\omega}_1^2=\frac12\boldsymbol{\bullet}\frac{\boldsymbol{m}\boldsymbol{l}^2}{1\boldsymbol{2}}\dot{\boldsymbol{\theta}}_1^2=\frac{\boldsymbol{m}\boldsymbol{l}^2}{24}\dot{\boldsymbol{\theta}}_1^2
\end{equation}
Kinetic energy of connecting rod 1:
\begin{equation}
T_{1}=T_{1\mathrm{r}}+T_{1_{P}}=\frac{ml^{2}}{6}\dot{\theta}_{1}^{2}
\end{equation}
Link 2:
\begin{equation}
\label{eq:t2p}
\begin{aligned}
T_{2p} &= \frac{1}{2} m v_{2}^{2} \\
&= \frac{5 m l^{2}}{8} \dot{\theta}_{1}^{2} + \frac{m l^{2}}{8} \dot{\theta}_{2}^{2} + \frac{m l^{2}}{4} \dot{\theta}_{1} \dot{\theta}_{2} + \frac{m l^{2}}{2} \cos\theta_{2} \dot{\theta}_{1} \left(\dot{\theta}_{1} + \dot{\theta}_{2}\right)
\end{aligned}
\end{equation}

\begin{equation}
T_{2\mathrm{r}}=\frac12J\omega_2^2=\frac{ml^2}{24}\left(\dot{\theta}_1^2+\dot{\theta}_2^2+2\dot{\theta}_1\dot{\theta}_2\right)
\end{equation}

Kinetic energy of connecting rod 2:
\begin{equation}
\label{eq:t2_total}
\begin{aligned}
T_{2} &= T_{2p} + T_{2r} \\
&= \frac{2 m l^{2}}{3} \dot{\theta}_{1}^{2} + \frac{m l^{2}}{6} \dot{\theta}_{2}^{2} + \frac{m l^{2}}{3} \dot{\theta}_{1} \dot{\theta}_{2} + \frac{m l^{2}}{2} \cos\theta_{2} \dot{\theta}_{1} \left(\dot{\theta}_{1} + \dot{\theta}_{2}\right)
\end{aligned}
\end{equation}

The total kinetic energy of the robot:
\begin{equation}
\begin{aligned}&T=T_1+T_2\\&=\frac{5ml^2}6\dot{\theta}_1^2+\frac{ml^2}6\dot{\theta}_2^2+\frac{ml^2}3\dot{\theta}_1\dot{\theta}_2+\frac{ml^2}2\cos\theta_2\dot{\theta}_1\left(\dot{\theta}_1+\dot{\theta}_2\right)\end{aligned}
\end{equation}
The dynamic equation is:
\begin{equation}
\boldsymbol{\tau}=\begin{bmatrix}\boldsymbol{M}_{11}&\boldsymbol{M}_{12}\\\boldsymbol{M}_{21}&\boldsymbol{M}_{22}\end{bmatrix}\begin{bmatrix}\ddot{\boldsymbol{\theta}}_1\\\ddot{\boldsymbol{\theta}}_2\end{bmatrix}+\begin{bmatrix}\boldsymbol{M}_{11}&\boldsymbol{M}_{12}\\\boldsymbol{M}_{21}&\boldsymbol{M}_{22}\end{bmatrix}\begin{bmatrix}\dot{\boldsymbol{\theta}}_1^2\\\dot{\boldsymbol{\theta}}_2^2\end{bmatrix}+\begin{bmatrix}\boldsymbol{C}_1\\\boldsymbol{C}_2\end{bmatrix}+\begin{bmatrix}\boldsymbol{G}_1\\\boldsymbol{G}_2\end{bmatrix}
\end{equation}

We have omitted the study of the human upper limb hand, treating it as an extended rigid body of the forearm. When people carry heavy objects, they often utilize the object's inertial force to increase their ability to handle heavy loads. By using the object's inertial force, they enable the heavy load to continue through areas with weaker bearing capacity, finally reaching the target position. Mimicking the method humans use to apply skillful force in carrying heavy objects, this approach can be applied to the trajectory planning of a robotic arm to significantly enhance its load-bearing capacity.

\section{DATA COLLECTION}
High-precision data obtained through Mediapipe and EMG signal acquisition systems provide a reliable basis for extracting human motion patterns, which is essential for optimizing trajectory planning and achieving efficient energy utilization.
\subsection{Upper Limb Motion Detection Based on MediaPipe}
The angular velocity, velocity, and acceleration of each joint were filtered. The experiment was set up with the initial state being the natural hanging down of the arms, and the final state being when the dumbbell position is slightly higher than the shoulder marker point, at which point the elbow joint cannot continue to flex. Multiple repeated measurements were conducted for this movement state, resulting in the determination of the angles of the shoulder and elbow joints relative to the vertical direction of the weight during the dumbbell curl process, as well as the acceleration at the wrist end. The test results were also filtered, as shown in Figure 3.
\begin{figure}[htb]
      \centering
      \includegraphics[width=\columnwidth]{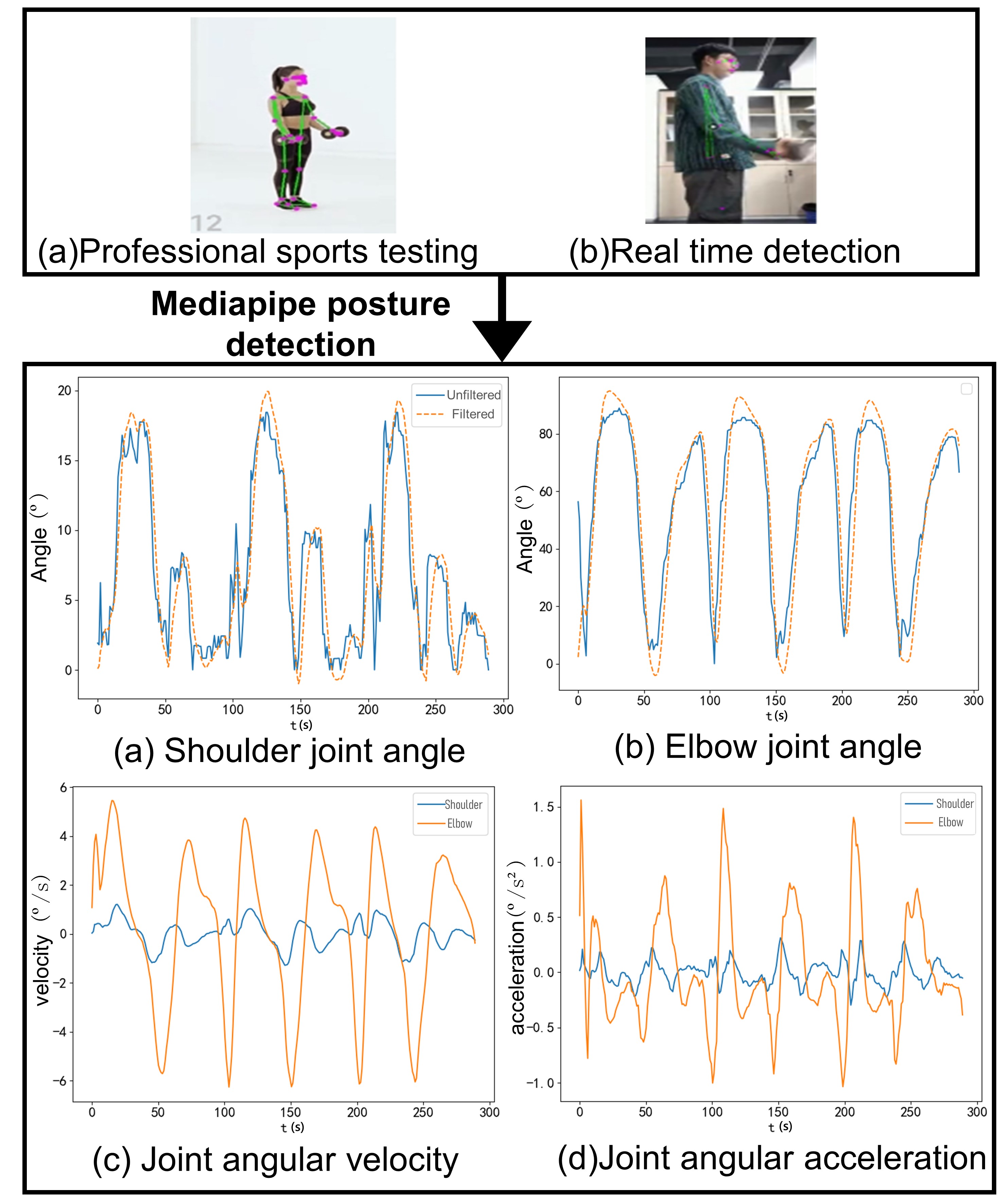}
      \caption{Motion posture detection based on mediapipe}
      \label{Figure 3}
   \end{figure}

According to the test results, it can be seen that the joint angles and angular velocities show a pattern of first increasing rapidly and then decreasing, with their peaks occurring in the first half of the motion.

\subsection{Upper Limb Movement Detection Based on Datalog EMG Signal Acquisition}
Utilizing a Datalog myoelectric acquisition device to collect contraction signals from different muscle groups of the human upper limb during the movement process, and analyzing the collected motion data. Myoelectric test point 1 is located on the deltoid muscle of the shoulder, test point 2 is on the triceps brachii, test point 3 is on the biceps brachii, and test point 4 is on the brevis muscle of the forearm near the elbow. The test load weight is 12 kg. After preparation, weightlifting movements are performed, and the acquired myoelectric signals are transmitted to the host computer via Bluetooth for data representation and graphical display.

\begin{figure}[htb]
      \centering
      \includegraphics[width=\columnwidth]{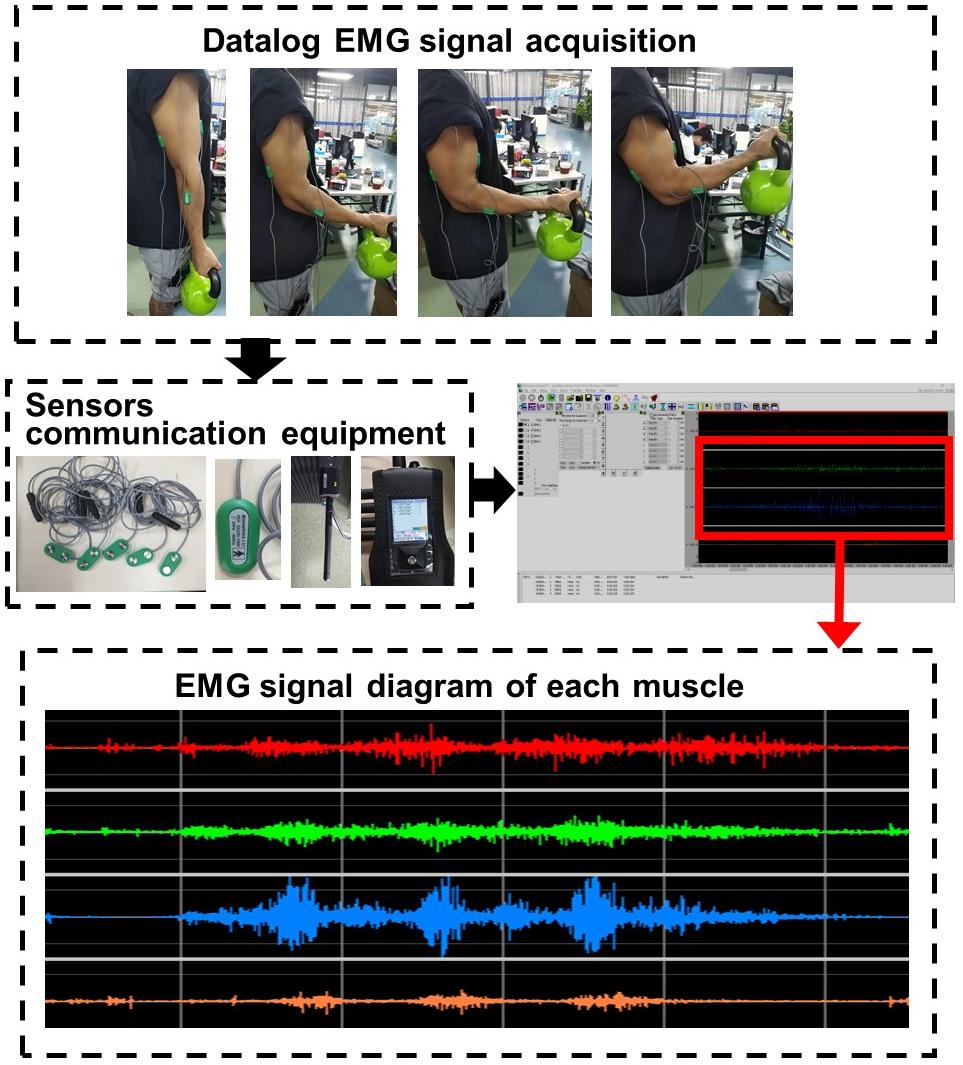}
      \caption{Collection of EMG signals for upper limb movements}
      \label{Figure 4}
   \end{figure}
From Figure 4, it can be observed that the entire process involved three lifting motions. Focusing on one of these motions for analysis, it can be seen that test point 1 on the deltoid muscle initiates the movement first, followed by the simultaneous engagement of the triceps and biceps muscles. Finally, the brevis muscle of the forearm near the elbow becomes engaged, with the weakest EMG signal. The biceps muscle exhibits the most intense changes, indicating that it bears the majority of the work. According to the trend of the EMG signals, it is evident that during the upward motion, the EMG signals show a parabolic distribution, indicating that the acceleration first increases and then decreases.

\section{TRAJECTORY PLANNING FOR ROBOTS}

\subsection{Selection of Predefined Trajectory Curves}

To enable the curtain wall installation robot to mimic the process of humans using skillful force in their work, we primarily consider the movement information of the shoulder and elbow joints related to the process of humans lifting a barbell with their upper limbs. To select typical curves, we compare trapezoidal curves, S-shaped curves, triangular curves, cubic polynomial curves, and quintic polynomial curves, as shown in Figure 5. Through this comparison, to ensure smooth and continuous curves during the movement process, we will adopt quintic polynomial interpolation for reasonable trajectory planning of the robot’s motion, taking into account the operating speed and acceleration of the robot during its movement.

\begin{figure}[htb]
      \centering
      \includegraphics[width=\columnwidth]{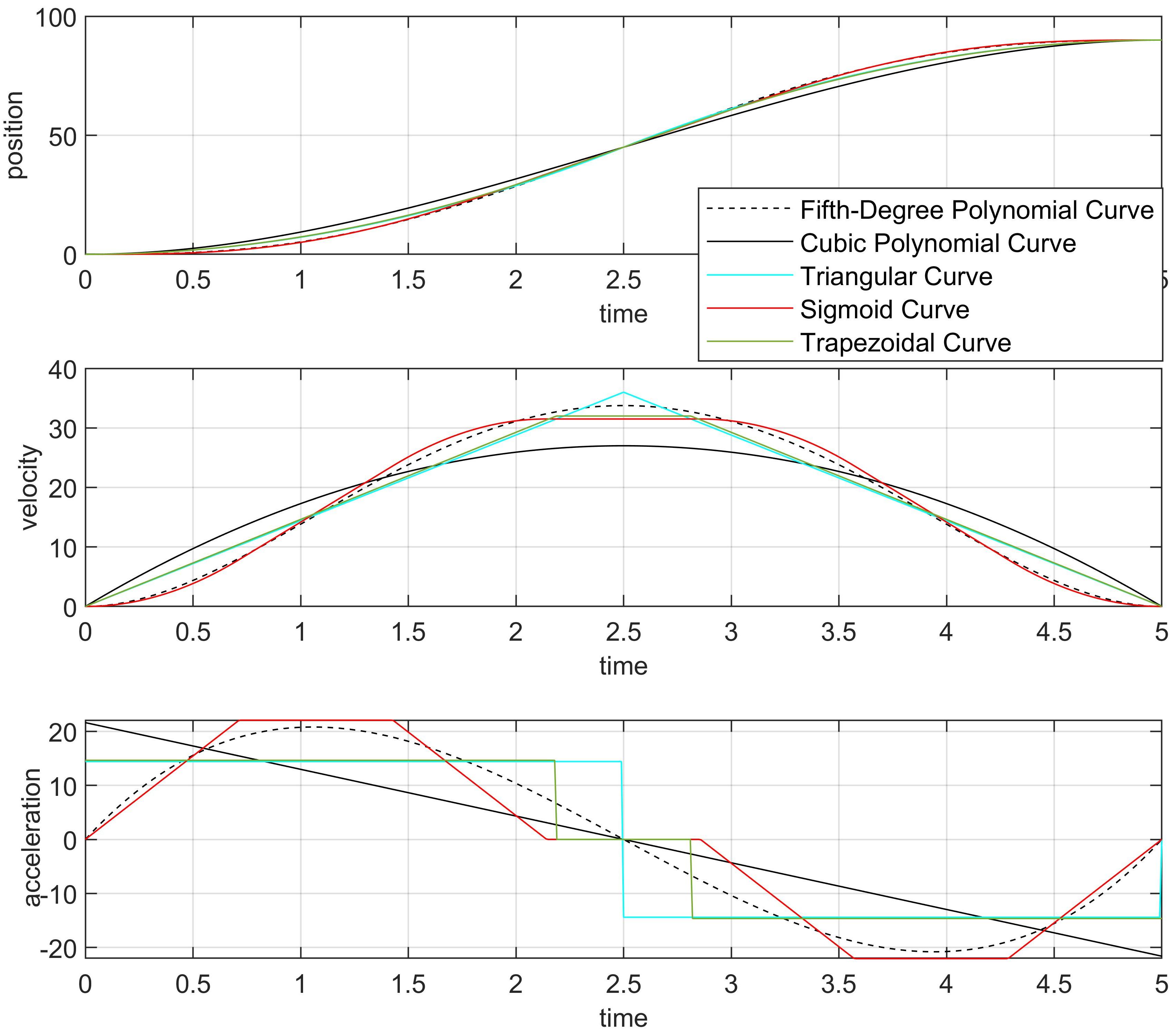}
      \caption{Comparison of velocity between different classical curves}
      \label{Figure 5}
   \end{figure}
The research on anthropomorphic motion mechanisms primarily considers the movement information of the shoulder and elbow joints related to the process of humans performing bicep curls with their upper limbs. We use quintic polynomial interpolation to reasonably plan the trajectory of the robot’s motion, which can effectively balance the operating speed and acceleration of the robot during its movement. The expression for the quintic polynomial function is:
\begin{equation}
\theta(t)=d_{0}+d_{1}t+d_{2}t^{2}+d_{3}t^{3}+d_{4}t^{4}+d_{3}t^{5}
\end{equation}
From the actual motion of the robotic arm, it can be observed that it stops during the entire motion process. It is known that the motion process starts and ends at specific positions, velocities, and accelerations, that is, the starting point \( \theta(t_0) \), \( \ddot{\theta}(t_{0}) \), the ending point \( \theta(t_f) \), \( \dot{\theta}(t_f) \), \( \ddot{\theta}(t_f) \), and the position \( \theta(t_i) \) information. For the fifth-order polynomial function.
\begin{equation}
\ddot{\theta}(t)=2d_{2}+6d_{3}t+12d_{4}t^{2}+20d_{3}t^{3}
\end{equation}
Substituting the known information of the start and end points, we get:
\begin{equation}
\label{eq:polynomial_motion}
\left\{
\begin{aligned}
\theta(t_{0}) &= d_{0} + d_{1} t_{0} + d_{2} t_{0}^{2} + d_{3} t_{0}^{3} + d_{4} t_{0}^{4} + d_{5} t_{0}^{5}, \\
\dot{\theta}(t_{0}) &= d_{1} + 2 d_{2} t_{0} + 3 d_{3} t_{0}^{2} + 4 d_{4} t_{0}^{3} + 5 d_{5} t_{0}^{4}, \\
\ddot{\theta}(t_{0}) &= 2 d_{2} + 6 d_{3} t_{0} + 12 d_{4} t_{0}^{2} + 20 d_{5} t_{0}^{3}, \\
\theta(t_{f}) &= d_{0} + d_{1} t_{f} + d_{2} t_{f}^{2} + d_{3} t_{f}^{3} + d_{4} t_{f}^{4} + d_{5} t_{f}^{5}, \\
\dot{\theta}(t_{f}) &= d_{1} + 2 d_{2} t_{f} + 3 d_{3} t_{f}^{2} + 4 d_{4} t_{f}^{3} + 5 d_{5} t_{f}^{4}, \\
\ddot{\theta}(t_{f}) &= 2 d_{2} + 6 d_{3} t_{f} + 12 d_{4} t_{f}^{2} + 20 d_{5} t_{f}^{3},
\end{aligned}
\right.
\end{equation}

Based on the conventional start and end point information, assuming the velocity and acceleration at the start and end points are 0, solving equation (11) ultimately yields the conventional motion trajectory, as shown in Figure 6.
\begin{figure}[htb]
      \centering
      \includegraphics[width=\columnwidth]{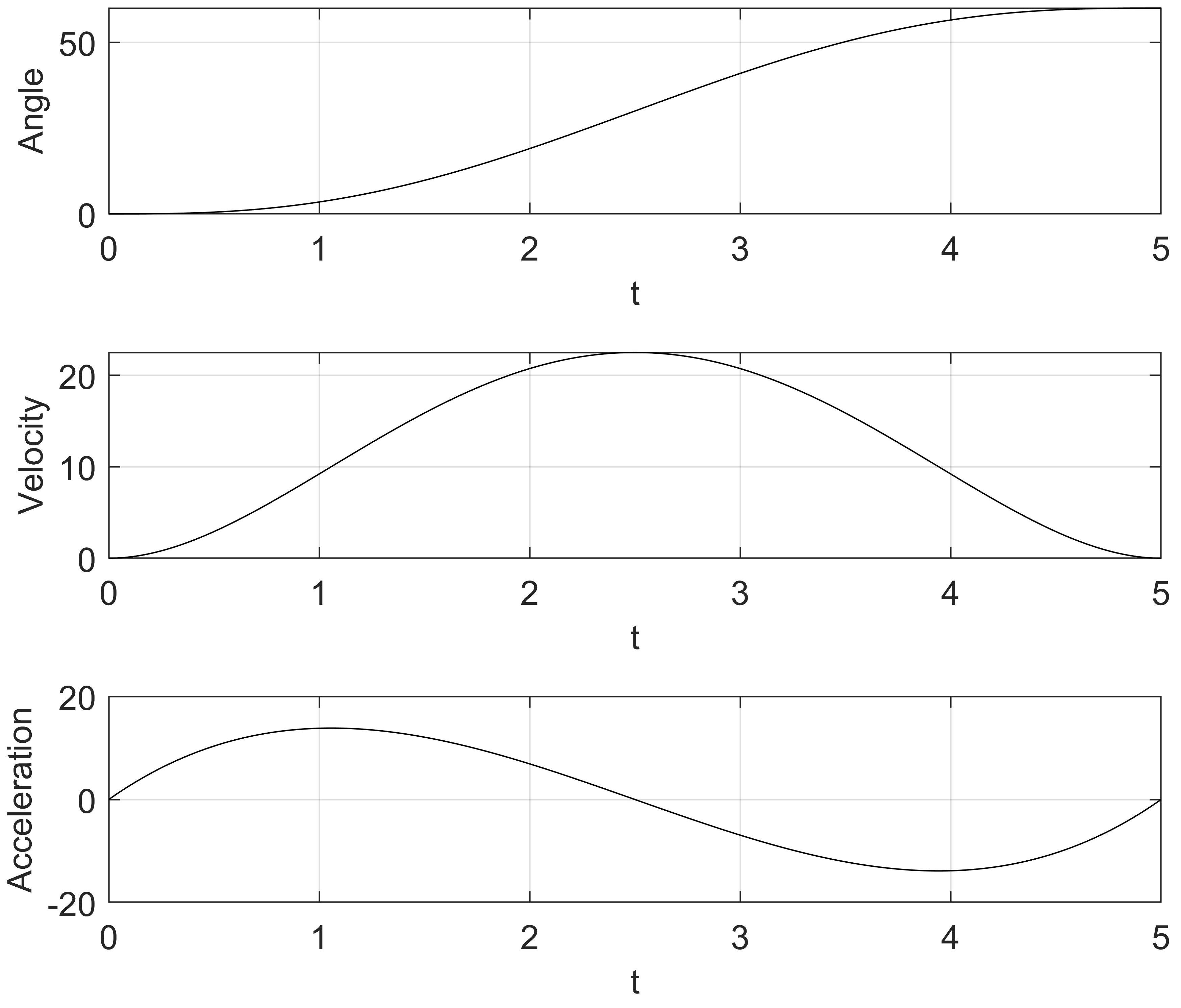}
      \caption{Fifth degree polynomial curve}
      \label{Figure 6}
   \end{figure}

\subsection{Bionic Mechanisms Based on Dataset}
Real-time EMG data from various muscle groups were collected using the Datalog EMG system. This enabled the identification of the activation sequence and intensity variations of key muscle groups during different movement phases, revealing force generation characteristics and sequencing. We could then accurately determine acceleration trends, identify key feature points, and provide a basis for optimizing acceleration and deceleration allocation.
We aligned EMG signal peaks with acceleration characteristic points to divide the movement into key phases: high-load phase, weakest load phase, and deceleration phase.

(A)High-load Capacity Region (e.g., low dumbbell position, elbow joint: 0°-40°): Short bursts of high acceleration rapidly increase kinetic energy, ensuring sufficient speed and energy before reaching the weakest load region.

(B)Weakest Load Region (elbow joint: 40°-90°): Kinetic energy is converted to potential energy to reduce joint torque, preventing excessive loads. This enables the arm to overcome its weakest load posture, enhancing load-bearing capacity.

(C)Deceleration Region (elbow joint: 90°-150°): Gravity or external resistance is used for deceleration, ensuring that the work done is primarily converted into potential energy changes for the task.

We use these characteristic points to optimize acceleration and deceleration allocation, enabling the robotic arm to overcome its weakest load posture. This ensures efficient mechanical energy transfer (kinetic + potential) throughout the movement cycle, providing crucial references for trajectory optimization.

\subsection{Trajectory Optimization Based on Particle Swarm Optimization}

Based on EMG and posture detection data, the movement process was divided into three phases: (A) High-load capacity phase, (B) Weakest load phase, and (C) Deceleration phase. Using the particle swarm optimization (PSO) algorithm, the velocity peak was placed at the center of phase (B)(approximately 62° at the elbow joint) to maximize kinetic energy throughout this phase, enabling the highest velocity through the weakest load-bearing posture. 

We applied the PSO algorithm to optimize fifth-order polynomial trajectories, with constraints set at a peak elbow velocity of 62°. The algorithm aimed to match key human motion characteristics in acceleration and velocity profiles, enhancing energy efficiency and load capacity. The optimized trajectory is compared with conventional trajectories in Figure 7.

\begin{figure}[htb]
      \centering
      \includegraphics[width=\columnwidth]{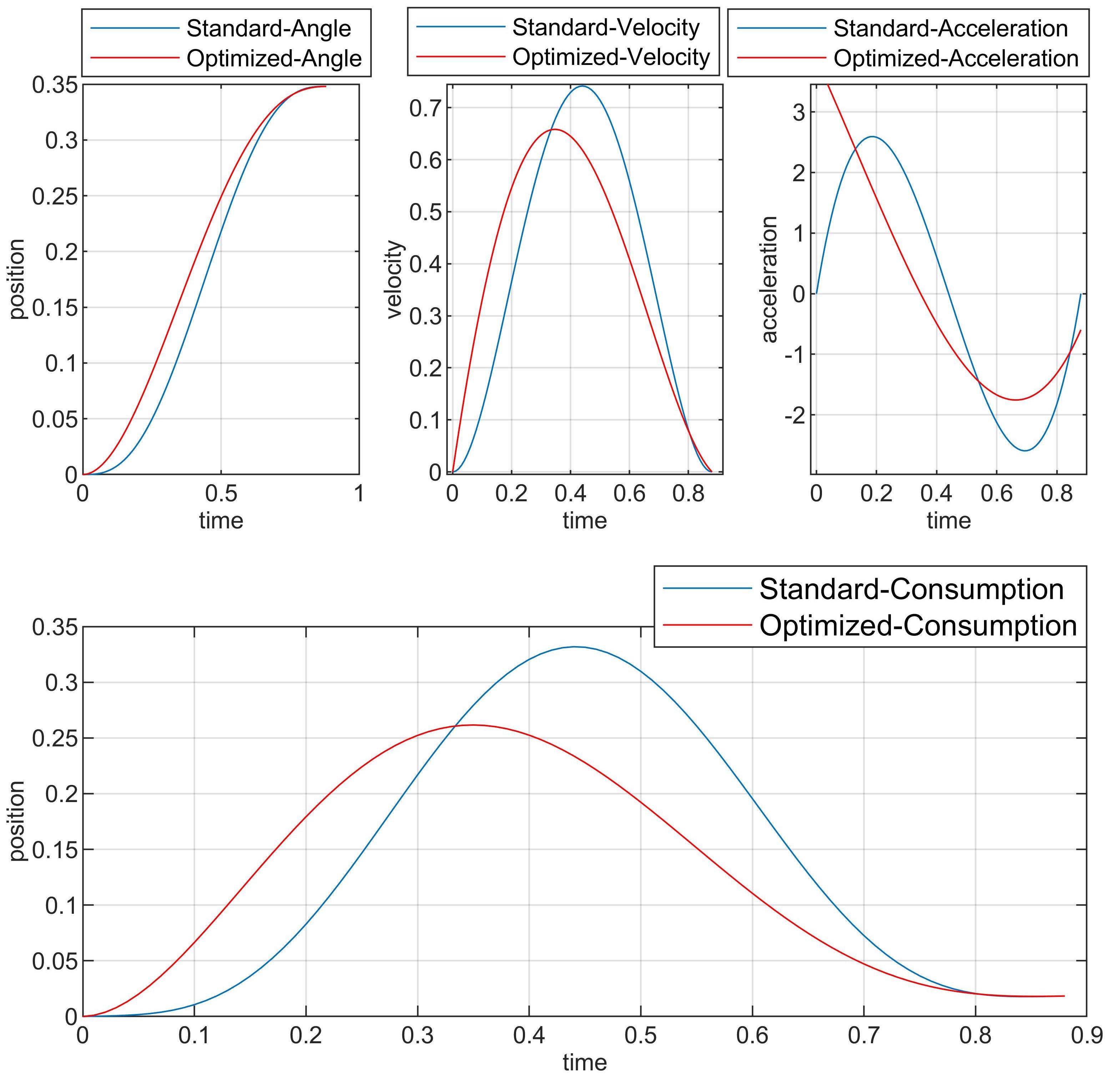}
      \caption{Comparison between Conventional Trajectory and Humanoid Optimization Trajectory}
      \label{Figure 7}
   \end{figure}
The Standard trajectory refers to the conventional trajectory, while the Optimized trajectory is the improved version. By comparing the energy consumption curves before and after planning, we have confirmed the effectiveness of this method: the optimized trajectory significantly reduces energy consumption by about 12\%.

\section{SIMULATION}
We establish a model of a curtain wall installation robot and validate the trajectory planning curves from the previous section using the ADAMS dynamics simulation software. To apply the movement characteristics of human upper limb bicep curls with a dumbbell, we choose a 6-degree-of-freedom curtain wall installation robot as the subject of our study and analyze it using the aforementioned motion control strategy. The forearm of this robotic arm is 0.45 m long, and the upper arm is 0.495 m long, with a 4 kg weight suspended at the end. The mechanism mapping diagram is shown in Figure 8.
\begin{figure}[htb]
      \centering
      \includegraphics[width=\columnwidth]{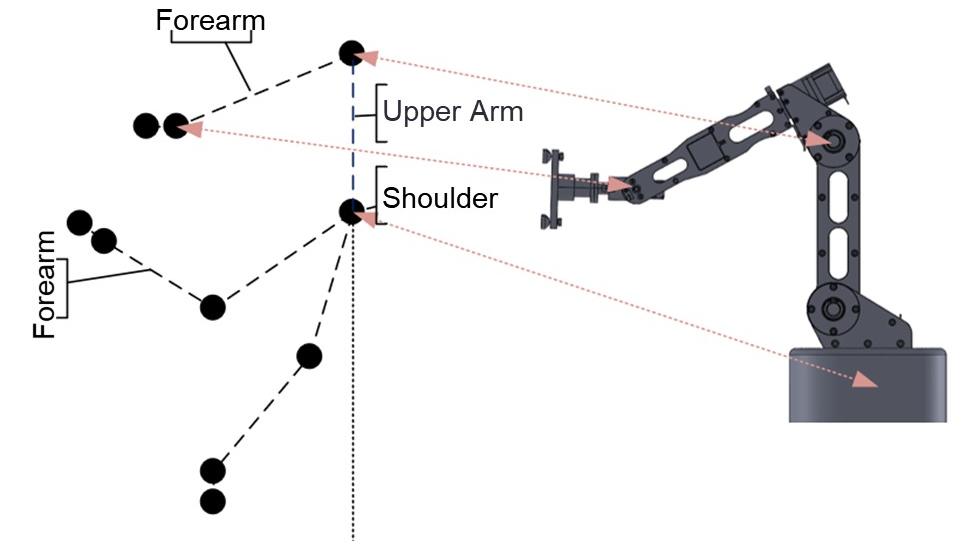}
      \caption{Mechanism mapping between human upper limbs and robots}
      \label{Figure 8}
   \end{figure}

In the mechanism mapping diagram of the human upper limb model and the robotic arm, the shoulder and elbow joints of the upper limb are mapped to the key rotational axes of the robotic arm. The shoulder joint of the upper limb is mapped to the base joint of the robotic arm, responsible for simulating the rotation and elevation movements of the shoulder. The elbow joint of the upper limb is mapped to the intermediate joint of the robotic arm, which is responsible for simulating the flexion and extension movements of the forearm. This mapping ensures that the robotic arm can reproduce the multi-degree-of-freedom movements of the human upper limb during dumbbell curls. The robotic arm achieves a similar range of motion and flexibility, providing biomechanical reference for executing complex tasks.
\begin{figure}[htb]
      \centering
      \includegraphics[width=\columnwidth]{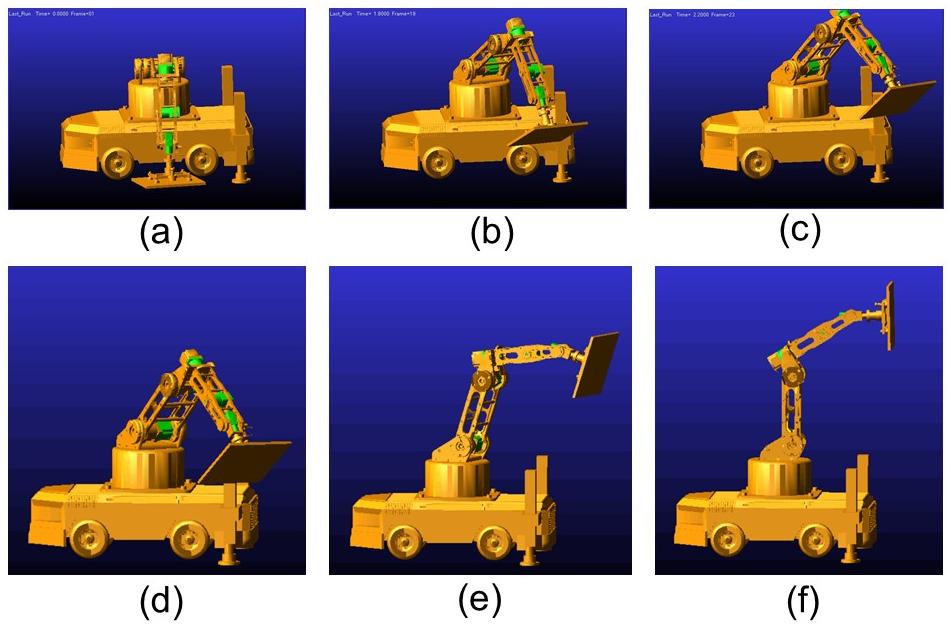}
      \caption{Simulation process diagram of curtain wall installation robot}
      \label{Figure 9}
   \end{figure}
\begin{figure}[htb]
      \centering
      \includegraphics[width=\columnwidth]{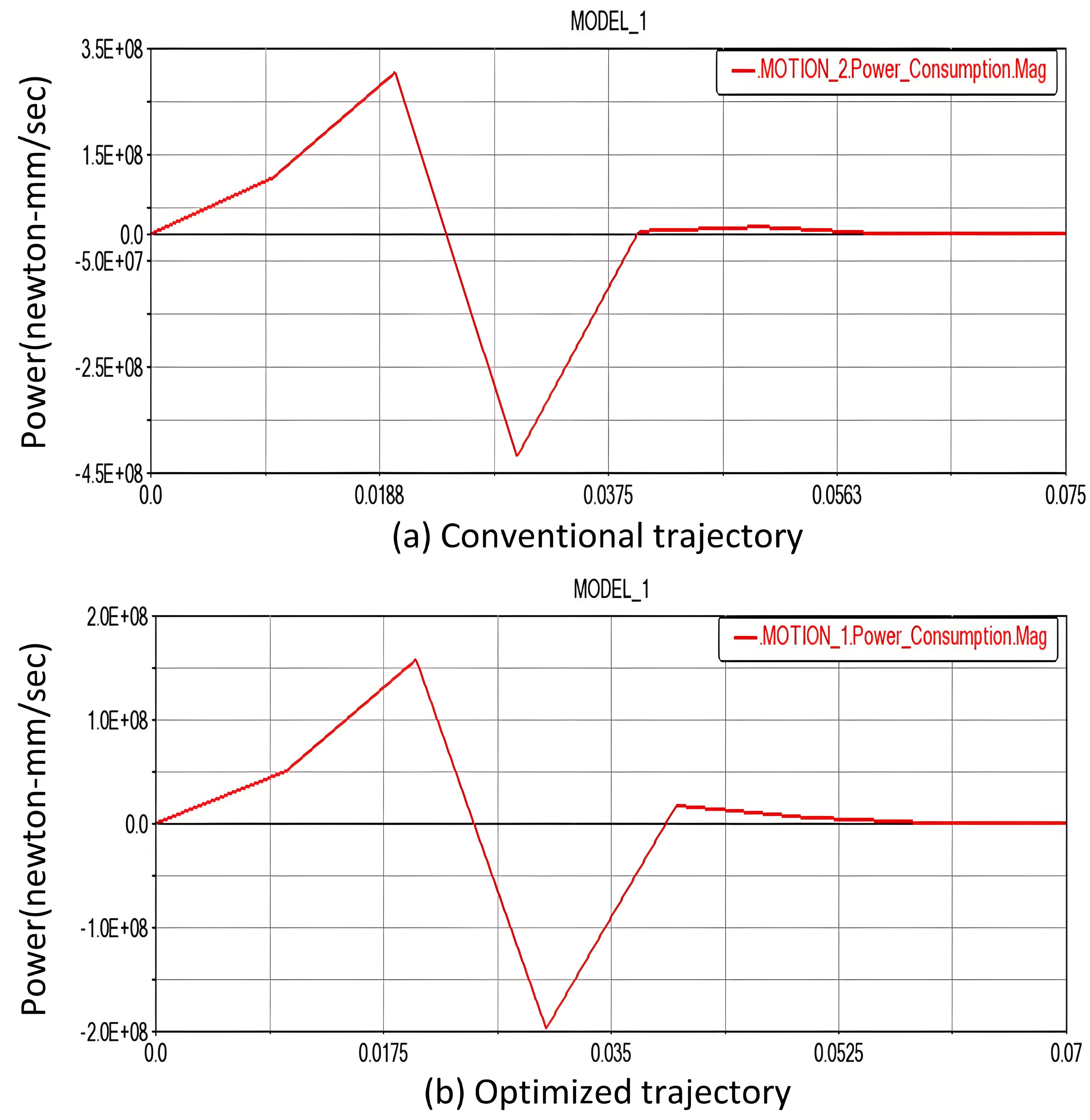}
      \caption{Simulation result diagram of curtain wall installation robot}
      \label{Figure 10}
   \end{figure}

Through simulation analysis of pre-optimization and post-optimization results, we demonstrated that incorporating characteristic human motion patterns into the trajectory significantly reduces energy consumption. The peak of the energy consumption curve decreased by 48.4\%. This confirms the superiority and practical effectiveness of the humanoid trajectory planning method, providing strong support for robotic arm applications such as curtain wall installation.

\section{CONCLUSIONS}

This study proposes an innovative trajectory planning method inspired by the biomechanics of human upper limb movements. We collected EMG signals during dumbbell curl exercises using surface EMG sensors (Datalog) and combined this data with posture detection from Mediapipe to establish a comprehensive motion-EMG coupling analysis framework. By analyzing energy conversion patterns, we identified key characteristic points in the movement. These points were integrated into the motion trajectory using Particle Swarm Optimization (PSO). The proposed trajectory planning method was validated through simulation experiments in a curtain wall installation scenario, demonstrating its superior performance.

In future work, more human motion experiments will be conducted to further establish the relationship between surface electromyography (EMG) signals, kinematic data, and energy optimization objectives. The current research has been validated solely through simulation. In the next stage, hardware experiments on physical robotic platforms will be conducted, and professional energy consumption analyzers will be utilized to perform real-world tests. These efforts aim to comprehensively evaluate the energy efficiency of the system and further validate and refine the methodology proposed in this study.

\addtolength{\textheight}{-5cm}   


\section*{ACKNOWLEDGMENT}

This work was supported in part by the National Key R\&D Program of China (No.2023YFB4705002), in part by the National Natural Science Foundation of China(U20A20283), in part by the Guangdong Provincial Key Laboratory of Construction Robotics and Intelligent Construction (2022KSYS 013), in part by the CAS Science and Technology Service Network Plan (STS) - Dongguan Special Project (Grant No. 20211600200062), in part by the Science and Technology Cooperation Project of Chinese Academy of Sciences in Hubei Province Construction 2023.


\end{document}